%% file: ETDAbs.tex
\documentclass{article}

\usepackage[final]{nips_2016}
\input{symbols}

\usepackage[utf8]{inputenc} 
\usepackage[T1]{fontenc}    
\usepackage{hyperref}       
\usepackage{url}            
\usepackage{booktabs}       
\usepackage{nicefrac}       
\usepackage{microtype}      
\usepackage{framed}
\usepackage{framed}
\usepackage{graphicx}
\usepackage[labelfont=bf]{caption}
\usepackage{subcaption}
\usepackage{tikz}
\usetikzlibrary{arrows,automata}
\usepackage{float}
\usepackage{bm,bbm,amsmath,amsfonts,amssymb}

\title{A First Empirical Study of \\ Emphatic Temporal Difference Learning}

\author{
Sina Ghiassian\\
Department of Computing Science\\
University of Alberta\\
\texttt{ghiassia@ualberta.ca} \\
\And
Banafsheh Rafiee\\
Department of Computing Science\\
University of Alberta\\
\texttt{rafiee@ualberta.ca} \\
\And
Richard S. Sutton\\
Department of Computing Science\\
University of Alberta\\
\texttt{rsutton@ualberta.ca} \\
}

\begin{document}
\def\tr{^\top\!}
\def\ph{{\bm\phi}}
\def\th{{\bm\theta}}
\renewcommand{\EE}[2]{\mathbb{E}_{#1\!\!}\left[#2\right]}
\newcommand{\EEb}[2]{\mathbb{E}_{#1\!}\bigl[#2\bigr]}
\newcommand{\hatEE}[2]{\hat{\mathbb{E}}_{#1\!\!}\left[#2\right]}
\newcommand{\EEpi}[1]{\EE{\pi}{#1}}
\newcommand{\EEmu}[1]{\EE{\mu}{#1}}
\newcommand{\CEE}[3]{\EE{#1}{{#2}~\middle\vert~{#3}}}
\renewcommand{\CEE}[3]{\EE{#1}{{#2}\mid{#3}}}
\def\CE#1#2{\CEE{\,}{#1}{#2}}
\def\CEpi#1#2{\CEE{\pi}{#1}{#2}}
\def\CEpip#1#2{\CEE{\pi'}{#1}{#2}}
\def\CEpistar#1#2{\CEE{\pi_*}{#1}{#2}}
\def\hatCE#1#2{\hatEE{\!}{{#1}\mid{#2}}}
\def\E#1{\EE{\,}{#1}}
\def\Epi#1{\EE{\pi}{#1}}
\def\Emu#1{\EE{\mu}{#1}}
\def\Epip#1{\EE{\pi'}{#1}}
\def\Ea#1{\EE{a}{#1}}
\def\CP#1#2{\Pr{#1\mid#2}}
\def\Pr#1{{{\rm Pr\!}\left\{#1\right\}}}
\def\Ppi#1{{{\rm Pr}_\pi\!}\left\{#1\right\}}
\def\Epib#1{\EEb{\pi}{#1}}

\maketitle

\begin{abstract}
In this paper we present the first empirical study of the emphatic temporal-difference learning algorithm (ETD), comparing it with conventional temporal-difference learning, in particular, with linear TD(0), on on-policy and off-policy variations of the Mountain Car problem. The initial motivation for developing ETD was that it has good convergence properties under \emph{off}-policy training (Sutton, Mahmood \& White 2016), but it is also a new algorithm for the \emph{on}-policy case.
In both our on-policy and off-policy experiments, we found that each method converged to a characteristic asymptotic level of error, with ETD better than TD(0).
TD(0) achieved a still lower error level temporarily before falling back to its higher asymptote, whereas ETD never showed this kind of ``bounce''. In the off-policy case (in which TD(0) is not guaranteed to converge), ETD was significantly slower.
\end{abstract}

\section{Emphatic Temporal Difference Learning}
We consider the problem of learning the value function for a Markov decision process and a given policy. An agent and environment interact at discrete time steps, $t=0, 1, 2, \ldots$,  at each of which the environment is in a state $S_t$, the agent selects an action $A_t$ and as a result the environment emits a reward $R_{t+1}$ and a next state $S_{t+1}$. States are represented to the agent as feature vectors $\ph_t = \ph(S_t) \in \Re^n$. We seek to find a parameter vector, $\th_t\in \Re^n$ such that the inner product $\th_t\tr\ph_t$ approximates the expected return $\CE{R_{t+1} + \gamma R_{t+2} + \gamma^2 R_{t+3} + \cdots}{A_{t:\infty}\sim\pi}$, where $\pi:\A\times\S\ra[0, 1]$ is a policy for selecting the future actions. In fact, all actions are selected by an alternate policy $\mu$. If $\pi=\mu$, then the training is called \emph{on-policy}, whereas if the two policies are different the training is called \emph{off-policy}.

We consider the special case of the emphatic temporal difference learning algorithm (ETD) in which bootstrapping is complete ($\lambda(s)=0, \forall s$) and there is no discounting ($\gamma(s)=1, \forall s$). Studying TD and ETD methods with complete bootstrapping is suitable because in this case the differences between them are maximized. As $\lambda$ approaches 1, the methods behave more similarly up to the point where they become equivalent when $\lambda=1$. By setting $\lambda=0$ and $\gamma=1$, the ETD algorithm can be completely described by:
\begin{align*}
\th_{t+1} &\doteq \th_{t} + \alpha\rho_{t}F_t \left(R_{t+1} + \th_{t}^{T}\ph_{t+1} - \th_{t}^{T}\ph_{t}\right) \ph_{t},\\
F_{t} &\doteq \rho_{t-1}F_{t-1} + 1, \textnormal{    with    } F_{0} \doteq 1,\\
\rho_t &\doteq \frac{\pi(A_t|S_t)}{\mu(A_t|S_t)},
\end{align*}
where $\alpha>0$ is a step size parameter. $F$ is the followon trace according to which the update at each time step is emphasized or de-emphasized. TD is obtained by removing the $F$ from the first equation. Because of $F$, ETD is different from TD even in the on-policy case in which $\rho$ is always 1. For a thorough explanation of ETD see (Sutton, Mahmood \& White 2016).

\section{Stability of On-policy TD with Variable $\lambda$: A Counterexample}

In this section we show that although the initial motivation for developing ETD was that it has good convergence properties under off-policy training (Yu 2015), it is also a different  algorithm under on-policy training. To emphasize the difference between the two, we present a simple example for which TD($\lambda$) is \emph{not convergent} under \emph{on-policy} training but ETD is. 

It has long been known that TD($\lambda$) converges with any constant value of $\lambda$ under on-policy training (Tsitsiklis \& Van Roy 1997). Surprisingly, TD($\lambda$) is not assured to converge with varying $\lambda$ even under on-policy training. Yu has recently presented a counterexample (personal communication) with state dependent $\lambda$ for which on-policy TD($\lambda$) is not convergent. The example is a simple Markov decision process consisting of two states in which the system simply moves from one state to another in a cycle. The process starts in each of the states with equal probability. Let $\lambda(S_1)=0$ and $\lambda(S_2)=1$, $\ph(S_1)=(3, 1)$ and $\ph(S_2)=(1, 1)$ and $\gamma=0.95$. As shown below, the TD($\lambda$) key matrix for this problem is not positive definite. Moreover, both eigenvalues of the key matrix have negative real parts and thus TD($\lambda$) diverges in this case.
\begin{center}
\begin{minipage}{.3\textwidth}
\begin{tikzpicture}[->,>=stealth',shorten >=1pt,auto,node distance=2.8cm,
                    semithick]
  \tikzstyle{every state}=[thick, draw=black,text=black]
  \node[state](A)                    {$S_1$};
  \node[state](C) [ right of=A] {$S_2$};
  \path (A) edge  [bend left]            node {} (C)
           (C) edge  [bend left]             node {} (A);
\end{tikzpicture}
\end{minipage}
\hspace{10mm}
\begin{minipage}{.3\textwidth}
$$
\textnormal{Key matrix} =
\begin{pmatrix}
      -0.4862  &  0.1713 \\
   -0.7787   &  0.0738
\end{pmatrix}
$$
\end{minipage}
\end{center}

This is while ETD is convergent under both off-policy and on-policy training with variable $\lambda$. This example appears in more detail in the supplementary material.

\section{Fixed-policy Mountain Car Testbed}

For our experimental study, we used a new variation of the mountain car control problem (Sutton \& Barto 1998) to form a prediction problem. The original mountain car problem has a 2-dimensional space, position (between -1.2 and 0.6), and velocity (between -0.07 and 0.07) with three actions, full throttle forward, full throttle backward, and 0 throttle. Each episode starts around the bottom of a hill (a uniform random number between -0.4 and -0.6). The reward is -1 on all time steps until the car pasts its goal at the top of the hill, which ends the episode. The task is undiscounted. Our variation of the mountain car problem has a fixed target policy which is to always push towards the direction of the velocity and not to push in any direction when the velocity is 0. We call the new variation of the mountain car problem, the fixed-policy mountain car testbed.

The performance measure we used is an estimation of the mean squared value error (MSVE) which reflects the mean squared difference between the true value function and the estimated value function, weighted according to how often each state is visited in the state space following the behavior policy: 

$$\widehat{MSVE}(\th)  = \frac{1}{|\mathcal{S}|} \sum_{s \in \mathcal{S}} [\hat{v}(s, \th) - v_{\pi}(s)]^2$$

$\mathcal{S}$ included 500 sample states gathered by following the behaviour policy for 10,000,000 steps and randomly choosing 500 states from the last 5,000,000. We did not use the first 5,000,000 because the state distribution may change as more steps are taken and the stationary distribution is achieved in the limit. The agent started from each state $s\in \mathcal{S}$ and followed the target policy to termination 1,000 times, each time the return was computed and recorded. All 1,000 returns were averaged and the result was used as the true value of the state value function, $v_{\pi}(s)$. The learning algorithm's estimation of the value function for state $s$ is shown by $\hat{v}(s, \th)=\th\tr\ph(s)$.

\section{On-Policy Experiments}
\label{sct:OnPolicyLearning}

We applied on-policy TD and on-policy ETD methods to the fixed-policy mountain car testbed. We created many instances of each method by changing the step size parameter. To approximate a value function for this problem, we used tile coding (Sutton 1996) with 5 tilings, 4$\times$4 tiles each. Each algorithm instance was initialized with a 0 weight vector, and then run for 500,000 episodes. The whole process was repeated for 50 runs.

\begin{figure*}[h!]
\begin{center}
\begin{subfigure}[t]{0.47\textwidth}
    \includegraphics[width = \textwidth]{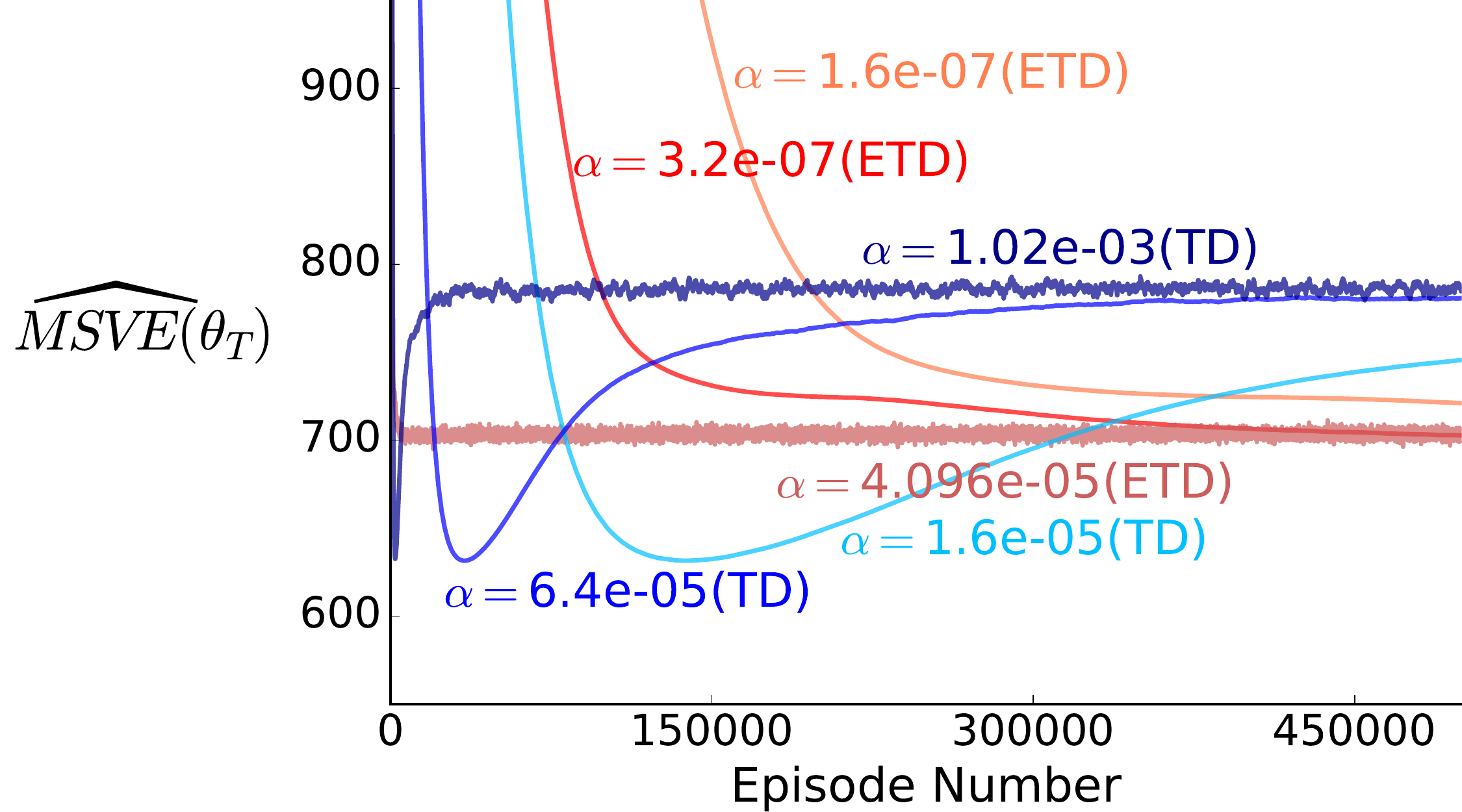}
    \caption{\textbf{Learning curves}: $\widehat{MSVE}$ was computed after each episode when the terminal state ($T$) was reached and thus is termed $\widehat{MSVE}(\th_T)$.}
    \label{fig:OnPolicyOverAlphaLearningCurves}
\end{subfigure}
\quad
\begin{subfigure}[t]{0.47\textwidth}
    \includegraphics[width = \textwidth]{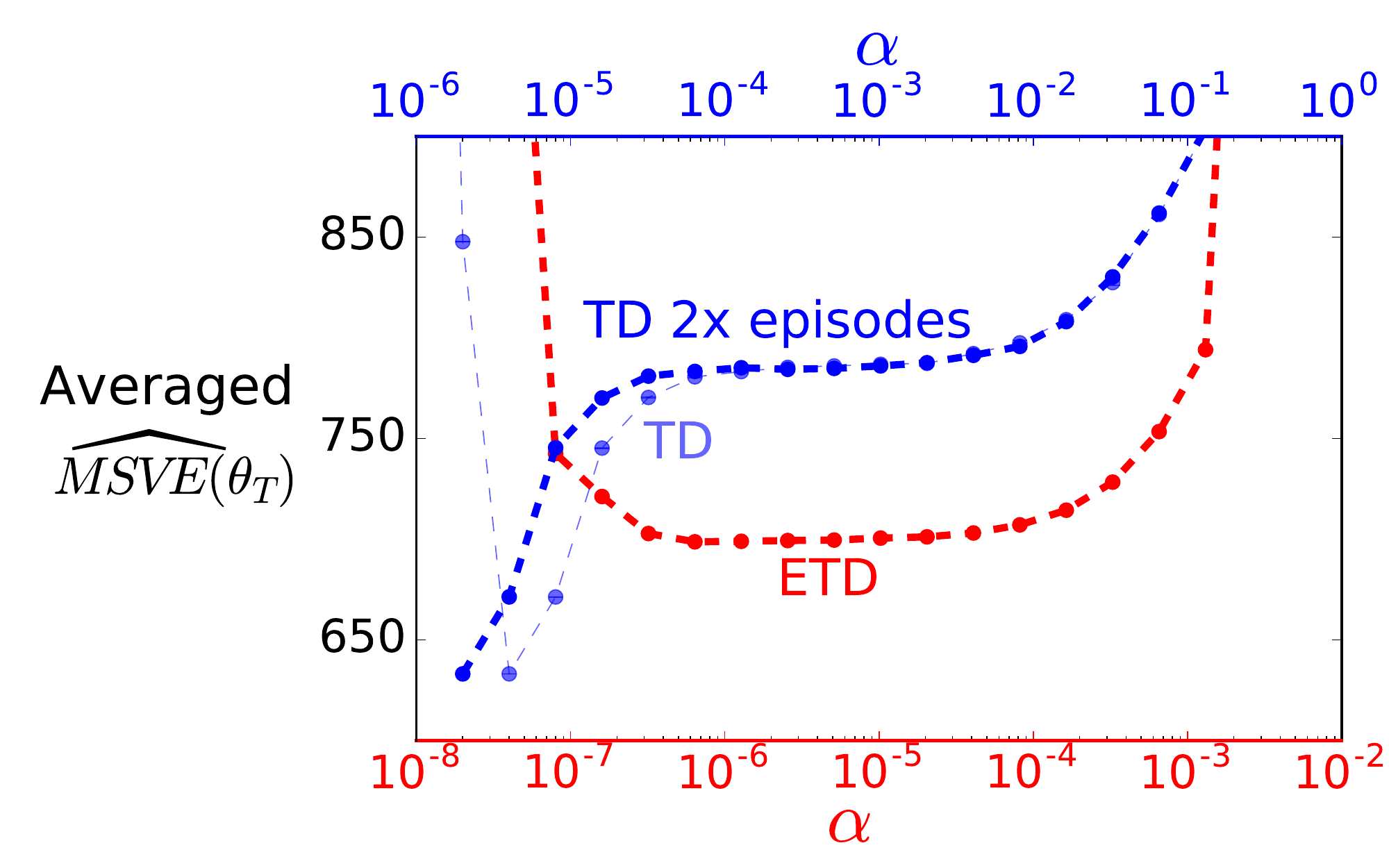}
    \caption{\textbf{Parameter study}: $\widehat{MSVE}$ was computed after each episode when the terminal state ($T$) was reached and then was averaged over the last 1\% episodes. This was done for each different value of $\alpha$. The standard errors of different instances are shown as error bars on the figure but are not visible due to their small sizes.}
    \label{fig:OnPolicyOverAlphaParameterStudy}
\end{subfigure}
\end{center}
\caption[]{{Results of on-policy ETD and TD methods on the fixed-policy mountain car testbed }}{}
\label{fig:OnPolicyOverAlpha}
\end{figure*}

To produce learning curves for each instance of the two methods we computed the error measure at the end of each episode and averaged over runs. See Figure \ref{fig:OnPolicyOverAlphaLearningCurves}. We also performed a parameter study of the asymptotic performance for both methods. To do so, we averaged the error of the last 1\% of the episodes for each run, and then computed the average and standard error over all 50 runs. See Figure \ref{fig:OnPolicyOverAlphaParameterStudy}.

\vspace{5mm}
To compare the performance of on-policy TD and on-policy ETD, we first need to understand how their errors changed as the number of episodes increased. ETD's error was a decreasing function of the number of episodes for sufficiently small values of $\alpha$. However, TD showed a bounce, reaching a low error temporarily before falling back to its higher asymptotic error. The depth and the asymptotic level of the bounce did not depend on $\alpha$, but its duration did. The smaller the $\alpha$, the later the bounce and as a result, it took more than 500,000 episodes for TD to converge for smaller values of $\alpha$. See Figure \ref{fig:OnPolicyOverAlphaLearningCurves}.

ETD outperformed TD in terms of asymptotic performance. TD instances with smaller values of the step size ($\alpha<10^{-4}$) did not converge within 500,000 episodes. See Figure \ref{fig:OnPolicyOverAlphaParameterStudy}. To confirm that TD has not converged for smaller values of $\alpha$, we repeated the TD experiments for 1,000,000 episodes and computed the error measure. The error measure changed \emph{only} for the instances that did not converged within 500,000 episodes. The light and the dark blue curves in Figure \ref{fig:OnPolicyOverAlphaParameterStudy} show the performance of different instances of the TD method after 500,000 and 1,000,000 episodes respectively. It is obvious that TD instances with $\alpha<10^{-4}$ did not converge while the instances with larger values of the step size did.

\section{Off-Policy Experiments}

\begin{figure*}[h!]
\begin{center}
\begin{subfigure}[t]{0.47\textwidth}
    \includegraphics[width = \textwidth]{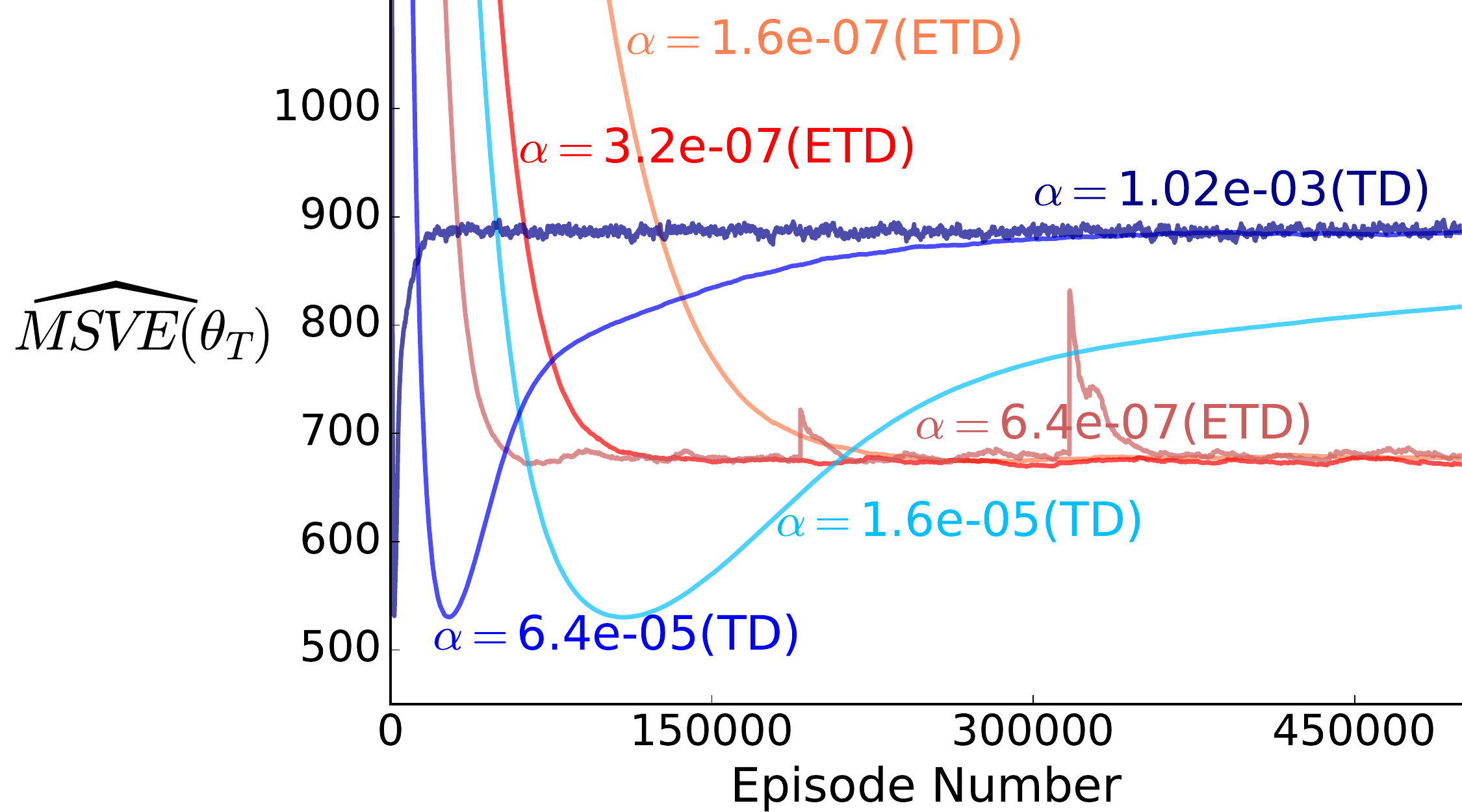}
    \caption{\textbf{Learning curves}: $\widehat{MSVE}$ was computed after each episode when the terminal state ($T$) was reached and thus is termed $\widehat{MSVE}(\th_T)$.}
    \label{fig:OffPolicyOverAlphaLearningCurves}
\end{subfigure}
\quad
\begin{subfigure}[t]{0.47\textwidth}
    \includegraphics[width = \textwidth]{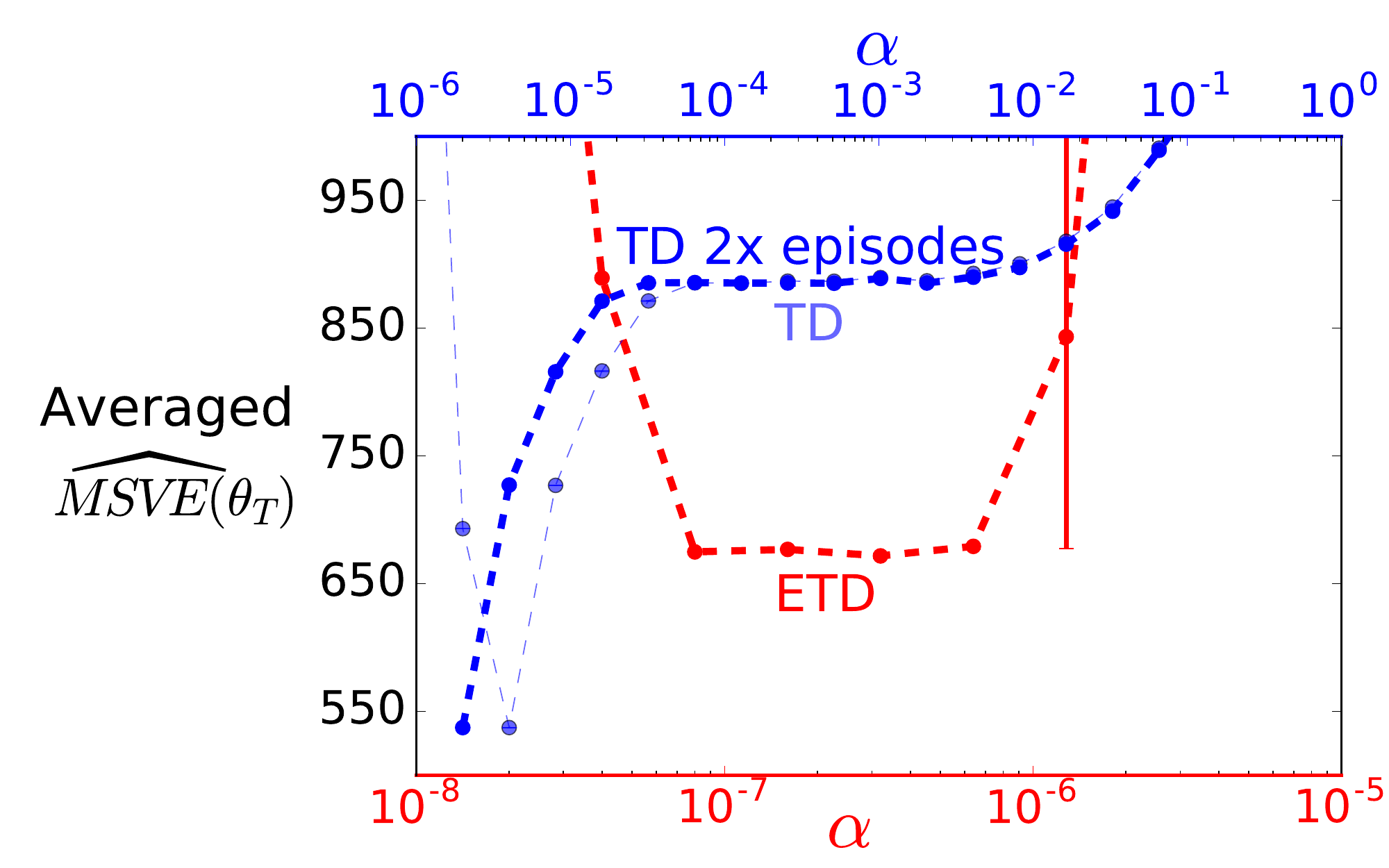}
    \caption{\textbf{Parameter study}: $\widehat{MSVE}$ was computed after each episode when the terminal state ($T$) was reached and then was averaged over the last 1\% episodes. This was done for each different value of $\alpha$. The standard errors of different instances are shown as error bars on the figure. Most of them are not visible due to their small sizes.}
    \label{fig:OffPolicyOverAlphaParameterStudy}
\end{subfigure}
\end{center}
\caption[]{{Results of off-policy ETD and TD methods on the fixed-policy mountain car testbed}}{}
\label{fig:OffPolicyOverAlpha}
\end{figure*}

We also applied off-policy TD and off-policy ETD to the fixed-policy mountain car testbed. In this case, the target policy was the same as the policy in the on-policy case and the behavior policy was to choose a random action 10\% of the time and act according to the target policy 90\% of the time. Again different instances of each method was created with different step size parameters. Each instance of the method was run for 500,000 episodes and the whole process was repeated for 50 runs. The learning curves for the off-policy case are presented in Figure \ref{fig:OffPolicyOverAlphaLearningCurves}. The parameter study results are in Figure \ref{fig:OffPolicyOverAlphaParameterStudy}.

\vspace{5mm}
Analogous to the on-policy case, each method had its advantages and disadvantages. ETD achieved a better asymptotic performance whenever it converged. This is while TD, compared to ETD, could take advantage of using larger values of step size and thus converged significantly faster (Figure \ref{fig:OffPolicyOverAlphaLearningCurves}).  ETD's step size values had to be set small (in the order of $10^{-7}$) to control the method's high variance (Figure \ref{fig:OffPolicyOverAlphaParameterStudy}). TD had a larger step size range with which it converged; however, ETD converged only for a short range of step size (Figure \ref{fig:OffPolicyOverAlphaParameterStudy}). Similar to the on-policy study, TD showed a bounce for every value of step size while ETD did not (Figure \ref{fig:OffPolicyOverAlphaLearningCurves}).

\section{Conclusion}

We performed the first systematic empirical study of the emphatic temporal difference learning method and showed that it can be used in a problem with a relatively large state space with promising results. Although ETD is originally proposed as an off-policy method, it can also be used as a reliable on-policy algorithm. According to our results, ETD seems to be slow in the off-policy case; however, it achieves a better asymptotic performance in both on-policy and off-policy cases. In spite of the fact that our experiments are limited to a variation of the mountain car problem, we believe that our observations can lead to a better understanding of both TD and ETD methods. Yu's counter example along with our experimental results motivate further study of ETD as an on-policy or off-policy method.

\subsubsection*{Acknowledgments}

The authors thank Huizhen Yu for insights and specifically for providing the counterexample. We gratefully acknowledge funding from Alberta Innovates Technology Futures and from the Natural Sciences and Engineering Research Council of Canada.

\section*{References}

Sutton, R.S. (1988). Learning to Predict by the Methods of Temporal Differences. Machine learning 3(1):9-44.

Sutton, R .S. (1996). Generalization in Reinforcement Learning: Successful Examples Using Sparse Coarse Coding. Advances in Neural Information Processing Systems:1038-1044.

Sutton, R. S., Barto, A. G. (1998). Reinforcement Learning: An Introduction. MIT Press.

Sutton, R. S., Mahmood, A. R., White, M. (2015). An Emphatic Approach to the Problem of Off-policy Temporal-difference Learning. The Journal of Machine Learning Research.

Tsitsiklis, J .N., Van Roy, B. (1997). An Analysis of Temporal-difference Learning with Function Approximation. IEEE Transactions on Automatic Control 42(5):674-690.

Yu, H. (2015). On convergence of emphatic temporal-difference learning. In Proceedings of the Conference on Computational Learning Theory.

\section{Supplementary Material}
\subsection{Stability of On-policy TD with Variable $\lambda$: A Counterexample}
Suppose the policy induces an irreducible Markov chain with transition matrix $P_\pi$ and a unique invariant probability distribution $\mu$ (i.e., $\mu^\top P_\pi = \mu^\top$).  
Let $D =diag(\mu)$ and let $\Phi$ be a feature matrix with linearly independent columns. 
The key matrix associated with the TD($\lambda$) algorithm is $A = \Phi^\top D (I - P_{\pi}^\lambda) \Phi$, where $P_\pi^\lambda$ is a substochastic matrix determined by $P_\pi$, $\lambda$ and the discount factor $\gamma$.
For a constant $\lambda \in [0,1]$, the matrix $A$ is positive definite (see e.g., Tsitsiklis and Van Roy 1997), ensuring the stability of the algorithm. 
This positive definiteness property relies critically on the fact that $\mu^\top P_{\pi}^\lambda < \mu^\top$, which does not hold in general when $\lambda$ is a function of states. Thus, with state-dependent $\lambda$, the positive definiteness of the matrix $A$ and the stability of the TD($\lambda$) algorithm are no longer guaranteed. 

In our example
$\lambda(S_1) = 0$, $\lambda(S_2) = 1$,
$\mu^\top=(0.5, 0.5)$ and
$P_\pi^\lambda = \left( \begin{matrix} 
      \gamma^2 & 0 \\
      \gamma & 0 
      \end{matrix} \right)$.
For $\gamma$ near $1$, e.g., $\gamma = 0.95$, and for $\Phi$ as given below, we can calculate the matrix $A$ associated with TD($\lambda$):
$$ \Phi = \left( \begin{matrix} 
      3 & 1 \\
      1 & 1 
      \end{matrix} \right), \qquad A = \Phi^\top D (I - P_{\pi}^\lambda) \Phi = \left( \begin{matrix} 
      -0.4862  &  0.1713 \\
   -0.7787   &  0.0738
      \end{matrix} \right).$$
The matrix $A$ is not positive definite. Moreover, both eigenvalues of $A$ have negative real parts, and hence $-A$ is not a Hurwitz matrix and TD($\lambda$) diverges in general in this case.    

\end{document}

%% file: symbols.tex
\newdimen\pHeight
\newdimen\pLower
\newdimen\pLineWidth
\newdimen\pKern
\newdimen\pIR
\newsavebox{\Cbox}
\newsavebox{\vertCmplx}
\newdimen\Cheight
\newdimen\Cwidth
\sbox{\Cbox}{\rm C}
\Cheight=\ht\Cbox
\Cwidth=\wd\Cbox
\advance\Cheight by \pHeight
\sbox{\vertCmplx}{\rule[\pLower]{\pLineWidth}{\Cheight}}
\sbox{\Cbox}{\usebox{\Cbox}\kern\pKern\usebox{\vertCmplx}}
\wd\Cbox=\Cwidth
\def\Re{\mathbb{R}}

\def\Nat{{\rm I\kern\pIR N}}

\newcommand{\EE}[1]{\exptE\left[#1\right]}

\def\A{{\mathcal{A}}}

\def\E{{\mathcal{E}}}

\def\S{{\mathcal{S}}}

\def\vec0{{\boldsymbol{0}}}

\newcommand{\ra}{\rightarrow}
\newcommand{\beq}{\begin{equation}}
\newcommand{\eeq}{\end{equation}}
\newcommand{\beqa}{\begin{eqnarray}}
\newcommand{\eeqa}{\end{eqnarray}}
\newcommand{\beqan}{\begin{eqnarray*}}
\newcommand{\eeqan}{\end{eqnarray*}}
\newcommand{\ben}{\begin{eqnarray*}}
\newcommand{\een}{\end{eqnarray*}}